\documentclass[10pt,twocolumn,letterpaper]{article}
\pdfoutput=1
\usepackage{cvpr}
\usepackage{times}
\usepackage{epsfig}
\usepackage{graphicx}
\usepackage{amsmath}
\usepackage{amssymb}
\usepackage{multirow}
\usepackage{cite}

\cvprfinalcopy 

\def\cvprPaperID{3565} 

\ifcvprfinal\fi
\begin{document}

\title{Attend More Times for Image Captioning}

\author{Jiajun Du\\
	Shanghai Jiao Tong University\\
	{\tt\small dujiajun@sjtu.edu.cn}
	\and
	Yu Qin\\
	Shanghai Jiao Tong University\\
	{\tt\small qinyu123@sjtu.edu.cn}
	\and
	Hongtao Lu\\
	Shanghai Jiao Tong University\\
	{\tt\small htlu@sjtu.edu.cn}
	\and
	Yonghua Zhang\\
	Bytedance\\
	{\tt\small mpeg21@hotmail.com}
}

\maketitle

\begin{abstract}

Most attention-based image captioning models attend to the image once per word. However, attending once per word is rigid and is easy to miss some information. Attending more times can adjust the attention position, find the missing information back and avoid generating the wrong word. In this paper, we show that attending more times per word can gain improvements in the image captioning task, without increasing the number of parameters. We propose a flexible two-LSTM merge model to make it convenient to encode more attentions than words. Our captioning model uses two LSTMs to encode the word sequence and the attention sequence respectively. The information of the two LSTMs and the image feature are combined to predict the next word. Experiments on the MSCOCO caption dataset show that our method outperforms the state-of-the-art. Using bottom up features and self-critical training method, our method gets BLEU-4, METEOR, ROUGE-L, CIDEr and SPICE scores of 0.381, 0.283, 0.580, 1.261 and 0.220 on the Karpathy test split.

\end{abstract}

\section{Introduction}

Image captioning is the task of generating a descriptive sentence for a given image. It is an important bridge to automatics image content understanding and connects two hot research fields, CV (computer vision) and NLP (natural language processing). Many image captioning models \cite{m-RNN}\cite{ShowTell}\cite{ShowAttendTell}\cite{BottomUp} have already been proposed but they still have a large gap in the caption quality like accuracy and fluency, compared with the captions given by human. MSCOCO image caption challenge \cite{COCOCaption} provides a large image captioning dataset to compare the performance of different image captioning models. On this platform, deep learning methods like Up-Down \cite{BottomUp} occupy the leaderboard.

Among the image captioning models, the attention-based encoder-decoder model \cite{ShowAttendTell} is most widely used. In this model, CNN (convolutional neural network) is used as the encoder to extract feature from the image. RNN (recurrent neural network) is used as the decoder to generate the caption. The image feature is used to initialize RNN's hidden state and RNN generates the caption word by word. In this process, attention mechanism is used to solve the bottleneck problem in the encoder-decoder architecture. In each time step, the model selectively attends to different regions of the image and generates an attention feature as the input of RNN.

For the conventional attention-based encoder-decoder model, the attention process is coupled with the word prediction process. The model attends to the image only once, before predicting the next word. However, the attention process of human is more complex. When a person want to give a caption for an image, he can attend to different regions of the image for many times before saying a word. It is easy to miss some information if attending to the image once per word. For example, the model might attend to the wrong place at the first glance. If the model can attend more times before generating the next word, it can adjust the attention place to the right place, find the useful information back and avoid generating the wrong word. 

\begin{figure}[t]
	\begin{center}
		\includegraphics[width=1\linewidth]{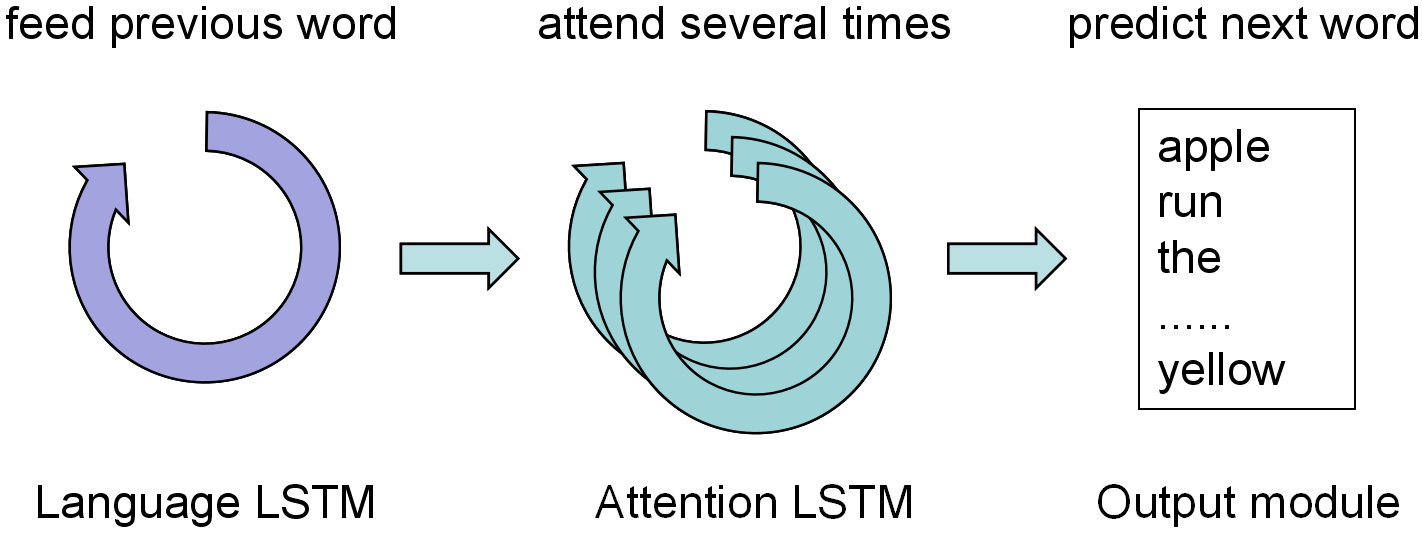}
	\end{center}
	\caption{Procedure of our image captioning model. First, feed the previous word to the language LSTM. Second, attend to the image and feed the attention feature to the attention LSTM. Repeat it for several times. Third, combine the image feature, the language LSTM state and the attention LSTM state to predict the next word.}
	\label{fig:Procedure}
\end{figure}

In this paper, we present a flexible two-LSTM merge model and show that attending more times per word can get higher caption scores than attending once. First, we introduce a two-LSTM merge model which decouples the encoding procedures of the word sequence and the attention sequence. The word sequence is encoded by the language LSTM. The attention sequence is encoded by the attention LSTM. The output module merges two LSTMs' hidden states and the average image feature to predict the next word. The conventional captioning model can only attend to the image once per word, but our two-LSTM merge model can attend for arbitary times and makes flexible attention processes possible.

Second, we also show that it is beneficial to attend more times per word in the image captioning task. Figure \ref{fig:Procedure} shows the procedure of attending more times per word. The first step is to feed the previous word to the language LSTM. The second step is to attend to the image, get the attention features and update the states of the attention LSTM for several times. The final step is to combine the information of the two LSTMs and the average image feature to predict the next word. We use the two-LSTM merge model to achieve the goal of attending more times per word and use this model to conduct experiments for comparing the effect of attending more times with attending once.

We evaluate our method on the MSCOCO caption dataset. The experiment results show that attending more times has higher scores than attending once in the image captioning task, especially for the CIDEr score. Besides, our best model outperforms previous state-of-the-art models on the MSCOCO caption dataset, achieving BLEU-4, METEOR, ROUGE-L, CIDEr and SPICE scores of 0.381, 0.283, 0.580, 1.261 and 0.220 respectively on the Karpathy test split.

\section{Related Work}

\subsection{Encoder-decoder architecture}

There are many architectures for image captioning. Mao et al. \cite{m-RNN} is the first one to use CNN to extract image features and use RNN to generate sentences in the image captioning task. They used RNN to generate the caption word by word and fed the image feature into RNN in every time step. Vinyals et al. \cite{ShowTell} proposed an encoder-decoder architecture and found it better to feed the image feature into RNN only in the first time step. Instead of feeding the image feature into RNN, Xu et al. \cite{ShowAttendTell} used the image feature to initialize the hidden state and cell state of LSTM \cite{LSTM} (long short-term memory). From then on, encoder-decoder architectures are widely used in the image captioning task and most image captioning models \cite{ShowTell}\cite{ShowAttendTell}\cite{VisualSentinel}\cite{SelfCritical}\cite{BottomUp} are a kind of encoder-decoder architectures.

\subsection{Merge model}

The merge model is a captioning model different from the encoder-decoder architecture and is first proposed by Hendricks et al. \cite{DCC}. They do not feed the image feature to the LSTM so that the LSTM only encodes the previous words and can be pretrained on large scale text corpus. Instead of only using the LSTM to predict the next word, they used a fully connected layer to combine the hidden state of the language LSTM and the image feature to predict the next word. Tanti et al. \cite{MergeModel} compared the merge model with the encoder-decoder architecture and found that the merge model is better than the encoder-decoder architecture in terms of caption quality and diversity. However, they did not apply attention mechanism to the merge model. In this paper, we not only apply attention mechanism to the merge model, but also add another LSTM to encode the attention sequence.

\subsection{Attention Mechanism}

Attention mechanism greatly improves many image captioning models' performance. Xu et al. \cite{ShowAttendTell} first used attention mechanism in image captioning and fed the attention feature into RNN in each time step. Lu et al. \cite{VisualSentinel} presented an opinion that it is not necessary to attend to the image during some words' prediction. They proposed a visual sentinel and used a gate to decide whether to attend to the image or the visual sentinel. These models attend to the image for at most one time per word. On the contrary, our model can attend to the image for several times and finally decide which word to use.

\subsection{Attention feature}

Xu et al. \cite{ShowAttendTell} used the last convolutional layer of ResNet \cite{ResNet} trained on ImageNet \cite{ImageNet} to generate attention features. Anderson et al. \cite{BottomUp} used object detection to get region features for each object candidate and made it easy to precisely attend to real objects. They used Faster R-CNN \cite{FasterRCNN} trained on Visual
Genome \cite{VisualGenome} to get the candidate object regions and extracted the region features by mean pooling the ResNet features over the target regions. Based on the detection features, Yao et al. \cite{VisualRelation} used a graph convolutional networks to encode the relationship of the objects into region features. In our work, detection features are used to conduct our experiments. However, more advanced attention features like Visual Relationship \cite{VisualRelation} can also be used in our method. The improvements of our method are orthogonal with the improvements of attention features. 

\section{Two-LSTM merge model}

\begin{figure}[t]
	\begin{center}
		\includegraphics[width=1\linewidth]{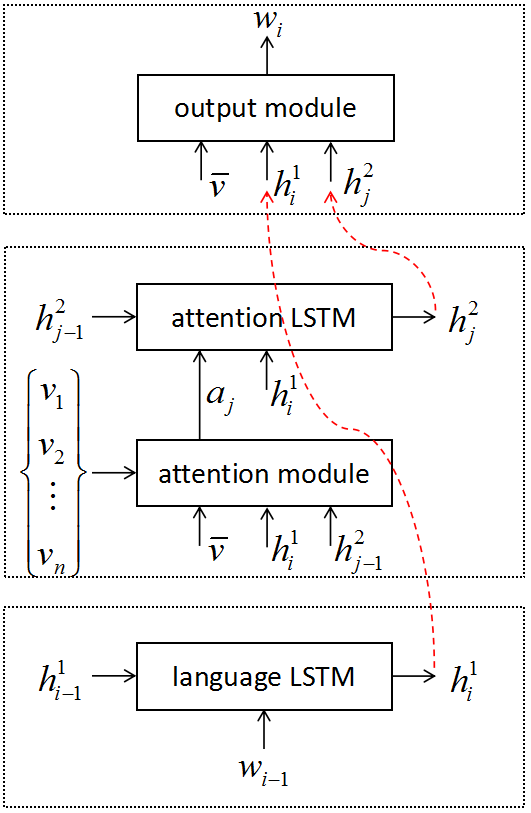}
	\end{center}
	\caption{Two-LSTM merge model. It consists of three parts, including the language LSTM, the attention part and the output module. Two LSTMs are used to encode the word sequence and the attention sequence respectively. The output module is used to combine the information of the image, the word sequence and the attention sequence to predict the next word. The two red dotted lines indicate the information flow from the language LSTM and the attention LSTM to the output module.}
	\label{fig:CaptionModel}
\end{figure}

Our captioning model is called two-LSTM merge model. As shown in Figure \ref{fig:CaptionModel}, our captioning model consists of three parts. The first part is the language LSTM. The second part includes the attention module and the attention LSTM. The third part is the output module. 

There are two LSTMs in our captioning model, including the language LSTM and the attention LSTM, which are used to encode the word sequence and the attention sequence respectively. By separating the encoding of the two sequences into two LSTMs, the two sequences can move forward individually. So our captioning model can attend more times before predicting the next word. In Figure \ref{fig:CaptionModel}, $i$ is the index of the word sequence and $j$ is the index of the attention sequence.  The two indice are different because the attention sequence moves faster than the word sequence when attending twice or more times per word. 

In fact, our model is a kind of merge models \cite{MergeModel}. The image features are not injected to the language LSTM. Only the attention LSTM and the output module take the image features as input. The attention LSTM takes weighted average image feature as input which is weighted by the probabilities predicted by the attention module. The output module takes average image feature as input. The average image feature and the hidden states of the two LSTMs are merged before sent to the output module. So our captioning model is called two-LSTM merge model.

\subsection{Language LSTM}

The language LSTM is used to encode previous generated words. In the $i$th iteration of the word sequence, the previous word $w_{i-1}$ is used to update the state of the language LSTM:

\begin{equation}
\label{LanguageLSTM}
h_i^1 = LSTM^1(w_{i-1}, h_{i-1}^1)
\end{equation}

\noindent
where $h_{i}^1$ refers to the hidden state and cell state of the language LSTM and $w_{i-1}$ is the word embedding \cite{Word2Vec} of the previous word. Word embedding maps each word to a low dimension vector. $W$ is the word embedding matrix and $w'$ is the one-hot enconding vector of the word. Then, the word embedding vector $w$ satisfies $w = Ww'$.

\subsection{Attention module and attention LSTM}

The attention module is used to generate the attention vector and the attention LSTM is used to encode the attention sequence. As shown in the middle part of Figure \ref{fig:CaptionModel}, we denote that the number of image features for the image is $n$, the $k$th image feature is $v_k$ and the average of them is $\overline{v}$.

In the $j$th iteration of the attention sequence, the attention module is used to generate the attention vector $a_j$. In the attention module, the $k$th image feature's attention weight $\alpha_{j,k}$ is generated by a two-layer network $f_{att}$. For the $k$th image feature $v_k$, the input of $f_{att}$ is $v_k$ together with the average image feature $\overline{v}$ and the hidden state $h_i^1$ and $h_{j-1}^2$ of the two LSTMs. Its hidden layer's activation function is the Tanh function. Its output layer's dimension is 1 so that we can get a scalar value $e_{j,k}$. Then, the softmax function is used to normalize $e_{j,k}$ into the attention weight $\alpha_{j,k}$.

\begin{equation}
\label{AttentionWeight}
e_{j,k} = f_{att}(v_k, \overline{v}, h_i^1, h_{j-1}^2)
\end{equation} 

\begin{equation}
\label{AttentionSoftmax}
\alpha_{j,k} = \frac{exp(e_{j,k})}{\sum_{t=1}^{n}exp(e_{j,t})}
\end{equation}

The attention vector $a_j$ is a weighted average of the image features $v_1, v_2, \dots v_{n}$. Each image feature $v_k$ is weighted by its attention weight $\alpha_{j,k}$.

\begin{equation}
\label{AttentionVector}
a_j = \sum_{k=1}^{n} \alpha_{j,k}v_k
\end{equation}

After getting the attention vector $a_j$, we use it to update the state of the attention LSTM. The attention LSTM takes the context vector $a_j$ and language LSTM's hidden state $h^1_i$ as input, together with the previous hidden state $h_{j-1}^2$, to update its state as:

\begin{equation}
\label{AttentionLSTM}
h_j^2 = LSTM^2(a_j, h_i^1, h_{j-1}^2)
\end{equation}

\subsection{Output module}

The output module is used to predict words. It combines the information of the image, the previous words and the attention sequence to predict the next word. The output module is also a two-layer network.  The activation function of its hidden layer is ReLU. The image feature $\overline{v}$, together with the hidden states of the two LSTMs $h^1_i$ and $h^2_j$, is concentrated to be the input of the output module. The softmax function is used to predict each word's probability. The conditional likelihood of the word given the image and previous words is:

\begin{equation}
\label{OutputModule}
p(w_i| w_1, w_2, \dots, w_{i-1}, image) = f_{out}(\overline{v}, h^1_i, h^2_j)
\end{equation}

\noindent
where $w_i$ is the $i$th word in the caption and $f_{out}$ is the output function implemented by the two-layer network and the softmax function.

\subsection{Inference procedure}

As shown in Figure \ref{fig:Procedure}, our model's inference procedure has three stages in each iteration of predicting words. First, the previous word is fed into the language LSTM. Then, our model attends to the image for several times. The state of the attention LSTM is updated by the attention features for several times as well. Finally, our model combines the information of the image, the language LSTM and the attention LSTM to predict the next word. In the validation process and test process, beam search is used to approximately get the caption with maximum conditional likelihood given the image.

\subsection{Training procedure}

In the training process, two kinds of training losses are experimented, including the cross entropy loss and the self-critical loss.

\subsubsection{Cross entropy loss}

The cross entropy loss is the negative log likelihood of the target word sequence given the image. $T$ is the number of words in the ground truth caption, $w_i^1$ is the $i$th word in the caption, and the cross entropy loss $L_1$ is:

\begin{equation}
\label{CrossEntropyLoss}
\begin{split}
L_1 &= - \log p(w_1^1, w_2^1, \dots, w_T^1 | image) \\
    &= - \sum_{i=1}^{T} \log p(w_i^1 | w_1^1, w_2^1, \dots, w_{i-1}^1, image)
\end{split}
\end{equation}

\noindent
where the conditional likelihood in each time step is predicted by the output module as Equation \ref{OutputModule}.

Scheduled sampling \cite{ScheduledSampling} is used in the training process to reduce the inconsistence between the training process and the inference process. With a certain probability, we sample the word according to the word probabilities predicted by the output module and feed the sampled word to the language LSTM, instead of the ground truth word.

\subsubsection{Self-critical loss}

The self-critical method is a policy gradient method which uses each sampled caption's CIDEr score $R$ as the reward. For the self-critical method, we do not feed the ground truth word to the language LSTM, but only feed the sampled word to it. To reduce the variance of the reward, the reward is subtracted by a baseline. The CIDEr score of the greedy decoding caption is used as the baseline $b$. $w_i^2$ is the $i$th word in the sampled caption and the self-critical loss $L_2$ is:

\begin{equation}
\label{SelfCriticalLoss}
L_2 = - (R - b)\sum_{i=1}^{T} \log p(w_i^2 | w_1^2, w_2^2, \dots, w_{i-1}^2, image)
\end{equation}

\subsection{Comparason with other models}

The original attention based image captioning models couple the encoding procedure of the attention sequence with the language sequence. Some of them use only one LSTM to encode the two sequence \cite{ShowAttendTell}. The others, in spite of having two LSTMs, only directly depend on one LSTM to predict the word \cite{BottomUp}. On the contrary, our language LSTM and attention LSTM are combined by our output module and can move forward individually, so our model is more flexible. 

\section{Experiments}

\subsection{Dataset}

We use MSCOCO caption dataset \cite{COCOCaption} to evaluate our method. MSCOCO caption dataset is composed of three parts, including the training set, the validation set and the test set. Each image in the training set and the validation set is annotated with 5 captions. However, there are no annotations for the test set. We use Karpathy data split \cite{KarpathySplit} to re-split the training set and the validation set into 113,287 training images, 5,000 validation images and 5,000 test images. Then, the test images have annotations and we calculate evaluation metrics for our test results according to their annotations and our model's prediction results. The evaluation metrics include BLEU \cite{BLEU}, METEOR \cite{METEOR}, ROUGE-L \cite{ROUGE}, CIDEr \cite{CIDEr} and SPICE \cite{SPICE}.

\subsection{Image features}

We extract bottom up features \cite{BottomUp} for images in the dataset. Bottom up features are generated by faster R-CNN \cite{FasterRCNN} pretrained on Visual Genome \cite{VisualGenome}. Faster R-CNN detects objects in the image and generates a list of bounding boxes together with their probabilities of containing specific objects. The objects with top 36 highest probabilities are chosen to generate the attention features. ResNet-101 \cite{ResNet} is used as the base network of Faster R-CNN. Each object's feature is generated by mean pooling over the corresponding region in the feature map of the last convolutional layer. Each image has 36 bottom up features and each image feature's dimension is 2048.

\subsection{Implementation details}

First, for captions with more than 16 words, we keep the beginning 16 words and remove the rest words. Then, we count each word's occurrence times in the captions and remove words which appear less than or equal to 5 times. The removed words are marked as the unknown words. After that, we get a word vocabulary of 9487, including the unknown word. The word embedding size is 512. The hidden sizes of the two LSTMs are both 1024. The hidden size of the attention module is 512. In the experiment, the model is trained by Adam \cite{ADAM} optimizer with batch size 16. Exponential learning rate decay is used. The learning rate is initially $5\times10^{-4}$ and is reduced by 20\% every 3 epochs. The final learning rate is about $3.17\times10^{-7}$. 

For the cross entropy loss training method, scheduled sampling is used. The scheduled sampling probability is initially 0\% and increases 5\% every 5 epochs before reaching 25\%. The scheduled sampling probability remains unchanged after reaching 25\%. For the self-critical training method, the model is first trained with cross entropy loss for 30 epochs and then continue to be trained with the self-critical method. We do not start self-critical training at the beginning because the word vocabulary is very large and it is hard to sample good captions from the beginning. After trained with the cross entropy loss, the model can sample good captions and can continue to be trained with the self-critical method.

For both the cross entropy loss training and the self-critical training, the model is trained for 100 epochs and evaluated on the validation set every 5,000 iterations. After each evaluation in the training process, the model is saved only if it gets a higher CIDEr score than before on the validation set. The model with the highest CIDEr score in the validation set is selected as the final model and evaluated on the test set. We use a beam size of 2 for beam search in both the validation process and the test process.

\subsection{Quantitative analysis}

\begin{table}
	\begin{center}
		\small
		\begin{tabular}{|l|c|c|c|c|c|c|}
			\hline
			N & M & BLEU-4 & METEOR & ROUGEL & CIDEr & SPICE\\
			\hline\hline
			\multirow{3}{*}{1} & 1 & 0.356 & 0.276 & 0.566 & 1.123 & 0.206\\
			& 2 & 0.353 & 0.273 & 0.562 & 1.120 & 0.205\\
			& 3 & 0.349 & 0.271 & 0.560 & 1.117 & 0.203\\
			\hline
			\multirow{3}{*}{2} & 1 & 0.343 & 0.270 & 0.558 & 1.087 & 0.200\\
			& 2 & \textbf{0.359} & 0.278 & 0.567 & 1.134 & 0.208\\
			& 3 & 0.358 & \textbf{0.279} & \textbf{0.568} & \textbf{1.139} & \textbf{0.209}\\
			\hline
			\multirow{3}{*}{3} & 1 & 0.322 & 0.261 & 0.548 & 1.024 & 0.193\\
			& 2 & 0.349 & 0.274 & 0.563 & 1.112 & 0.204\\
			& 3 & \textbf{0.359} & 0.277 & 0.566 & 1.129 & 0.206\\
			\hline
		\end{tabular}
	\end{center}
	\caption{Cross entropy loss training results on the validation set. N is the number of attention iterations in the training process. M is the number of attention iterations in the validation process.}
	\label{table:CrossEntropyValidation}
\end{table}

\begin{table}
	\begin{center}
		\small
		\begin{tabular}{|l|c|c|c|c|c|c|}
			\hline
			N & M & BLEU-4 & METEOR & ROUGEL & CIDEr & SPICE\\
			\hline\hline
			\multirow{3}{*}{1} & 1 & 0.376 & 0.283 & 0.581 & 1.248 & 0.217\\
			& 2 & 0.371 & 0.279 & 0.576 & 1.228 & 0.212\\
			& 3 & 0.367 & 0.277 & 0.574 & 1.218 & 0.212\\
			\hline
			\multirow{3}{*}{2} & 1 & 0.358 & 0.280 & 0.576 & 1.213 & 0.214\\
			& 2 & 0.377 & 0.283 & \textbf{0.582} & 1.252 & 0.217\\
			& 3 & 0.374 & 0.282 & 0.580 & 1.240 & 0.216\\
			\hline
			\multirow{3}{*}{3} & 1 & 0.340 & 0.273 & 0.560 & 1.134 & 0.206\\
			& 2 & 0.377 & 0.282 & 0.578 & 1.239 & 0.216\\
			& 3 & \textbf{0.382} & \textbf{0.284} & \textbf{0.582} & \textbf{1.258} & \textbf{0.219}\\
			\hline
		\end{tabular}
	\end{center}
	\caption{Self-critical training results on the validation set. N is the number of attention iterations in the training process. M is the number of attention iterations in the validation process.}
	\label{table:SelfCriticalValidation}
\end{table}

\begin{table}
	\begin{center}
		\small
		\begin{tabular}{|l|c|c|c|c|c|c|}
			\hline
			Attend & BLEU-4 & METEOR & ROUGEL & CIDEr & SPICE\\
			\hline\hline
			Once & 0.356 & 0.277 & 0.564 & 1.133 & 0.206\\
			More & \textbf{0.358} & \textbf{0.278} & \textbf{0.566} & \textbf{1.150} & \textbf{0.210}\\
			\hline
		\end{tabular}
	\end{center}
	\caption{Cross entropy loss training results on the test set. The first model attends once in the training and test process. The second model attends more times. The number of attention iterations is determined by the best result on the validation set.}
	\label{table:CrossEntropyTest}
\end{table}

\begin{table}
	\begin{center}
		\small
		\begin{tabular}{|l|c|c|c|c|c|c|}
			\hline
			Attend & BLEU-4 & METEOR & ROUGEL & CIDEr & SPICE\\
			\hline\hline
			Once & 0.375 & 0.282 & \textbf{0.580} & 1.252 & 0.216\\
			More & \textbf{0.381} & \textbf{0.283} & \textbf{0.580} & \textbf{1.261} & \textbf{0.220}\\
			\hline
		\end{tabular}
	\end{center}
	\caption{Self-critical training results on the test set. The first model attends once in the training and test process. The second model attends more times. The number of attention iterations is determined by the best result on the validation set.}
	\label{table:SelfCriticalTest}
\end{table}

In the experiments, we evaluate the influence of attention iterations per word on the generated captions. The evaluation metrics include BLEU-4, METEOR, ROUGE-L, CIDEr and SPICE. Different numbers of attention iterations in the training and validation (test) process are allowed.

\begin{table*}
	\begin{center}
		\begin{tabular}{|l|c|c|c|c|c|c|}
			\hline
			Method & BLEU-4 & METEOR & ROUGEL & CIDEr & SPICE\\
			\hline\hline
			m-RNN \cite{m-RNN} & 0.190 & 0.228 & - & 0.842 & -\\
			NIC \cite{ShowTell} & 0.183 & 0.237 & - & 0.855 & -\\
			Hard attention \cite{ShowAttendTell} & 0.250 & 0.230 & - & - & -\\
			Visual sentinel \cite{VisualSentinel} & 0.332 & 0.266 & - & 1.085 & -\\
			Self-critical \cite{SelfCritical} & 0.342 & 0.267 & 0.557 & 1.140 & -\\
			Regularize RNN \cite{RegularRNN} & 0.335 & 0.261 & 0.546 & 1.034 & 0.190\\
			Up-Down \cite{BottomUp} & 0.363 & 0.277 & 0.569 & 1.201 & 0.214\\
			\hline
			Ours & \textbf{0.381} & \textbf{0.283} & \textbf{0.580} & \textbf{1.261} & \textbf{0.220}\\
			\hline
		\end{tabular}
	\end{center}
	\caption{Single model performance comparison on the Karpathy test split of MSCOCO caption dataset. Our method has the highest scores compared with these previous methods.}
	\label{table:Comparison}
\end{table*}

\newcommand{\tabincell}[2]{\begin{tabular}{@{}#1@{}}#2\end{tabular}}

\begin{table*}
	\begin{center}
		\begin{tabular}{|c|p{2.7in}|p{2.7in}|}
			\hline
			Image & Our results & Ground truth\\
			\hline\hline
			\tabincell{c}{
				\includegraphics[width=1in, height=1in]{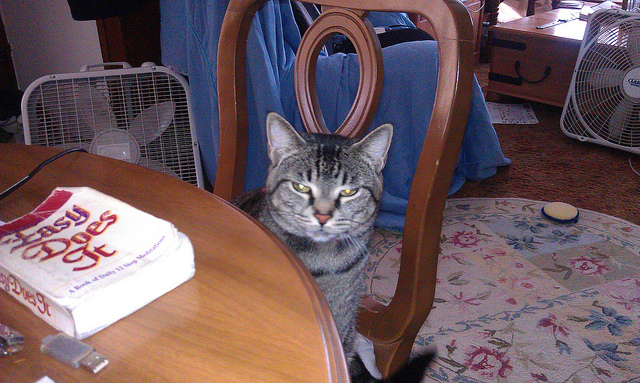} 
			} &
			\tabincell{l}{
				Once: a cat sitting on a chair in a room \\
				More: a cat sitting in a chair next to a table \\
			} &
			\tabincell{l}{
				1. A gray tiger cat sitting at a wooden table on a \\ chair \\
				2. A grey cat sitting in chair next to a table \\
				3. A cat sitting in a chair pulled up to a table \\
				4. A cat sitting in a chair at a table with a book on \\ it \\
				5. A cat sitting in a chair by a table \\ 
			} \\
			\hline
			\tabincell{c}{
				\includegraphics[width=1in, height=1in]{COCO_val2014_000000462565} 
			} &
			\tabincell{l}{
				Once: a man riding a bike down a street \\
				More: a group of people riding bikes down a \\ street \\
			} &
			\tabincell{l}{
				1. People on bicycles ride down a busy street \\
				2. A group of people are riding bikes down the \\ street in a bike lane \\
				3. Bike riders passing Burger King in city street \\
				4. A group of bicyclists are riding in the bike lane \\
				5. Bicyclists on a city street, most not using the \\ bike lane \\
			} \\
			\hline
			\tabincell{c}{
				\includegraphics[width=1in, height=1in]{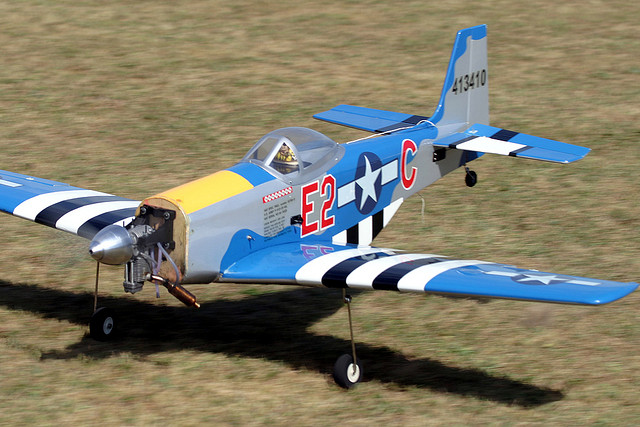} 
			} &
			\tabincell{l}{
				Once: a small airplane is on a grassy field \\
				More: a small blue and white airplane is in the \\ grass \\
			} &
			\tabincell{l}{
				1. A small blue plane sitting on top of a field \\
				2. An E2 airplane painted blue with black and \\ white stripes \\
				3. An old warplane is on display in a field \\
				4. A blue small plane standing at the airstrip \\
				5. Model airplane with an American insignia and \\ stripes on wings \\
			} \\
			\hline
			\tabincell{c}{
				\includegraphics[width=1in, height=1in]{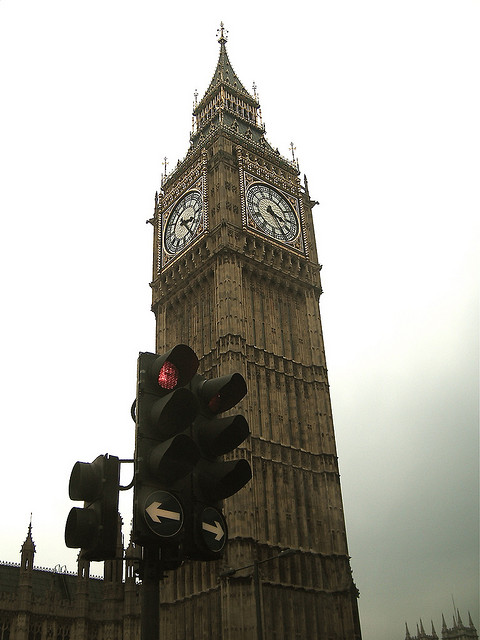} 
			} &
			\tabincell{l}{
				Once: a traffic light sitting next to a tall building \\
				More: a traffic light in front of a tall tower \\
			} &
			\tabincell{l}{
				1. An intersection that is close to a clock tower \\
				2. The Big Ben clock tower towering over a traffic \\ light \\
				3. A tower with a clock on it in front of a street \\ light \\
				4. A clock tower extends into the sky in a city \\
				5. A very tall tower with a big clock in the middle \\ of it \\
			} \\
			\hline
		\end{tabular}
	\end{center}
	\caption{Captions generated by our models. The images are chosen from the test set. 'Once' means attending once per word and 'More' means attending more times per word. For those images, captions generated by attending more times are more descriptive  than attending once.}
	\label{table:Examples}
\end{table*}

\subsubsection{Validation results}

Table \ref{table:CrossEntropyValidation} and \ref{table:SelfCriticalValidation} show the validation results for different attention iterations. $N$ is the number of attention iterations in the training process. $M$ is the number of attention iterations in the validation process. In Table \ref{table:CrossEntropyValidation}, the model is trained by minimizing the cross entropy loss. In Table \ref{table:SelfCriticalValidation}, the model is trained by the self-critical method. 

As shown in Table \ref{table:CrossEntropyValidation} and \ref{table:SelfCriticalValidation}, the number of attention iterations in the training and validaition process both has a strong influence on the validation scores. Generally speaking, it is better to make attention iterations $N$ and $M$ close to each other, because the inconsistence between the training process and the validation process is likely to degrade the performance of our model. When $N$ is equal to $M$, attending more times can always get higher CIDEr score. For the cross entropy loss training method, the CIDEr score is 1.123 for $N=M=1$, 1.134 for $N=M=2$ and 1.129 for $N=M=3$. For the self-critical training method, the CIDEr score is 1.248 for $N=M=1$, 1.252 for $N=M=2$ and 1.258 for $N=M=3$.

However, in the experiments, $N$ is not always equal to $M$ when we get the best results on the validation set. This is because there is some randomness for the training results. For the cross entropy loss training method, the highest CIDEr score is achieved by attending 2 times in the training process and then attending 3 times in the validation process. For the self-critical training method, the highest CIDEr score is achieved by attending 3 times both in the training process and in the validation process. All hyper parameters, except the number of attention iterations, are the same as described in section 4.3.

Attending more times improves the validation scores, especially for the CIDEr score. For cross entropy training, attending once gets BLEU-4, METEOR, ROUGE-L, CIDEr and SPICE scores of 0.356, 0.276, 0.566, 1.123 and 0.206 while attending more times (attend 2 times in the training process and attend 3 times in the validation process) gets 0.358, 0.279, 0.568, 1.139 and 0.209. For self-critical training, attending once gets BLEU-4, METEOR, ROUGE-L, CIDEr and SPICE scores of 0.376, 0.283, 0.581, 1.248 and 0.217 while attending more times (attend 3 times both in the training process and in the validation process) gets 0.382, 0.284, 0.582, 1.258 and 0.219.

\subsubsection{Test results}

Table \ref{table:CrossEntropyTest} and \ref{table:SelfCriticalTest} show the test results of attending once and attending more times. For attending more times, we use the model with the highest CIDEr score on the validation set. The number of attention iterations in the test process is determined by the number of attention iterations in the validation process when getting the highest CIDEr score. 

Table \ref{table:CrossEntropyTest} shows the cross entropy loss training results on the test set. For the cross entropy loss, attending more times uses the model which attends 2 times in the training process. It attends 3 times in the test process. Table \ref{table:SelfCriticalTest} shows the self-critical training results on the test set. For the self-critical loss, attending more times uses the model which attends 3 times in the training process. It also attends 3 times in the test process.

As shown in Table \ref{table:CrossEntropyTest} and \ref{table:SelfCriticalTest}, attending more times also improves the test scores, especially for the CIDEr score. For cross entropy training, attending once gets a CIDEr score of 1.133 while attending more times gets a CIDEr score of 1.150. For self-critical training, attending once gets a CIDEr score of 1.252 while attending more times gets a CIDEr score of 1.261.

\subsubsection{Comparison with state-of-the-arts}

As shown in Table \ref{table:Comparison}, we compare our method with m-RNN \cite{m-RNN}, NIC \cite{ShowTell}, hard attention \cite{ShowAttendTell}, visual sentinel \cite{VisualSentinel}, self-critical \cite{SelfCritical}, regularize RNN \cite{RegularRNN} and Up-Down \cite{BottomUp} on the Karpathy test split \cite{KarpathySplit}. Here, we do not compare our method with Visual Relationship \cite{VisualRelation} because their innovations are in the feature level which is orthogonal to our method and they use the relationship annotations of the external dataset Visual Genome \cite{VisualGenome} to train their model. The evaluation metrics include BLEU-4, METEOR, ROUGE-L, CIDEr and SPICE. We use our best single model which attends three times per word in both the training process and the test process and is trained by the self-critical method. Our model gets highest scores on all the five evaluation metrics, especially CIDEr. The previous state-of-the-art Up-Down \cite{BottomUp} has a CIDEr score of 1.201 but our method has a CIDEr score of 1.261.

\subsection{Qualitative analysis}

We give four examples in Table \ref{table:Examples} to show our method's effects. The four images are chosen from the test set. One caption is generated by attending once and the other is generated by attending more times. The two captioning models trained in Table \ref{table:SelfCriticalTest} are used to generate the two kinds of captions. Attending once is easy to miss some information and attending more times can find it back. For the top left image, attending once misses the existence of the table. For the top right image, attending once does not find that there are more people on the back of the man. For the bottom left image, attending more times adds the color of the airplane. For the bottom right image, attending once does not recognize what the tall building is, but attending more times recognizes that it is a tower. From those examples, we can find that attending more times can generate more descriptive captions.  

\section{Conclusion}

In conventional image captioning models, the number of attentions is equal to the number of words. In this paper, we present a flexible two-LSTM merge model to encode more attentions than words, which uses two LSTMs to encode the word sequence and the attention sequence respectively. By conducting experiments involving our captioning model, we show that attending more times can generate captions better than attending once per word in the image captioning task. Besides, attending more times does not increase the number of parameters in the captioning model, compared with attending once. Our best model achieves state-of-the-art performance on the MSCOCO caption dataset.

\section{Future work}

In the future, we want to add a controller to decide how many times the captioning model should attend before predicting the next word. Then, the model can attend different times for each word and is more flexible. For some word, the controller can even choose to not attend and just move forward the language LSTM to predict the next word.

{\small
\bibliographystyle{ieee}
\bibliography{attend_more}
}

\end{document}